%% file: main.tex
\documentclass{article}
\usepackage{amsmath,graphicx}

\usepackage{booktabs}
\usepackage{tabularx}
\usepackage{xcolor}

\usepackage[preprint]{spconf}
\copyrightnotice{\copyright\ IEEE 2023}
\toappear{To appear in {\it Proc.\ ICASSP2024,
   April 14-19, 2024, Seoul, Korea}}

\newcommand{\unpublished}{\cite{serdyukAudioVisualSpeechRecognition2021, shillingfordLargeScaleVisualSpeech2019, sonLipReadingProfile2017, makinoRecurrentNeuralNetwork2019}}

\def\presec{\vspace{-5pt}}
\def\postsec{\vspace{-5pt}}
\def\presub{\vspace{-10pt}}
\def\postsub{\vspace{-5pt}}

\title{LiteVSR: Efficient Visual Speech Recognition by Learning from Speech Representations of unlabeled data}
\name{Hendrik Laux$^{1}$, Emil Mededovic$^{1}$, Ahmed Hallawa$^{2}$, Lukas Martin$^{2}$, Arne Peine$^{2}$, Anke Schmeink$^{3}$}

\address{
$^1$University Hospital RWTH Aachen, Germany \quad
$^2$Clinomic Medical GmbH, Aachen, Germany\\
$^3$Chair of Information Theory and Data Analytics, RWTH Aachen University, Germany
}

\begin{document}
\ninept


\maketitle
%
\begin{abstract}
\presec
This paper proposes a novel, resource-efficient approach to Visual Speech Recognition (VSR) leveraging speech representations produced by any trained Automatic Speech Recognition (ASR) model. Moving away from the resource-intensive trends prevalent in recent literature, our method distills knowledge from a trained Conformer-based ASR model, achieving competitive performance on standard VSR benchmarks with significantly less resource utilization. Using unlabeled audio-visual data only, our baseline model achieves a word error rate (WER) of 47.4\% and 54.7\% on the LRS2 and LRS3 test benchmarks, respectively. After fine-tuning the model with limited labeled data, the word error rate reduces to 35\% (LRS2) and 45.7\% (LRS3). Our model can be trained on a single consumer-grade GPU within a few days and is capable of performing real-time end-to-end VSR on dated hardware, suggesting a path towards more accessible and resource-efficient VSR methodologies.
\end{abstract}


\section{Introduction}
\postsec
\label{sec:intro}

Visual Speech Recognition (VSR), commonly known as lip reading, describes the task of transcribing speech based purely on visual cues from lip movements. VSR proves instrumental in scenarios such as interpreting lip movements from speech-impaired patients~\cite{lauxTwostageVisualSpeech2023} and enhancing Automatic Speech Recognition (ASR) in acoustically challenging environments \cite{serdyukAudioVisualSpeechRecognition2021, xuDiscriminativeMultiModalitySpeech2020}. Despite sharing similarities with ASR, the inherent ambiguity in the visual domain, exemplified by the existence of visually indistinguishable characters or homophemes \cite{goldschenRationalePhonemevisemeMapping1996}, exacerbates the challenge of VSR for both human and machine lip-readers. The necessity to process high-dimensional video data further elevates the complexity and resource demand in training end-to-end VSR models. Recent progress in the field has enabled VSR systems based on deep neural networks to outperform professional human lip readers \cite{chungLipReadingSentences2017, shillingfordLargeScaleVisualSpeech2019} and significantly lower the error rates on common VSR benchmarks \cite{shiLearningAudioVisualSpeech2022, liuSynthVSRScalingVisual2023, prajwalSubwordLevelLip2022}. However, this often incurs substantial computational resource requirements and the creation of large audio-visual training corpora that remain unavailable to the public research community \unpublished. This indicates the need for new methods of training VSR models that can attain competitive performance with publicly available data and feasible computational resources. 

\presub
\subsection{Related Work}
\postsub

The first end-to-end deep neural network for sentence-level VSR, "LipNet" \cite{assaelLipNetEndtoEndSentencelevel2016}, was introduced in 2016, employing a combination of 3D convolutions and Gated Recurrent Units (GRU) trained using the Connectionist Temporal Classification (CTC) loss \cite{gravesConnectionistTemporalClassification2006}. 
While LipNet used the GRID audio-visual corpus \cite{cookeAudiovisualCorpusSpeech2006} for training, the Lip Reading Sentences (LRS) datasets quickly emerged as the primary benchmark for VSR research \cite{chungLipReadingSentences2017, afourasLRS3TEDLargescaleDataset2018}. In contrast to GRID, which comprises a relatively small number of videos recorded under controlled conditions, the LRS datasets present a much larger variety of challenging head poses, sentence structures, and environmental settings.

Given the natural similarity to speech recognition models, recent advancements in machine lip-reading have benefited from the adoption of architectures like Transformers \cite{shiLearningAudioVisualSpeech2022,vaswaniAttentionAllYou2017} and Conformers \cite{gulatiConformerConvolutionaugmentedTransformer2020, maEndToEndAudioVisualSpeech2021}. 
While earlier publications use the CTC training objective for aligning the input and output sequence \cite{shillingfordLargeScaleVisualSpeech2019, afourasASRAllYou2020}, recent works shift towards more resource-intensive sequence-to-sequence models with auto-regressive decoding schemes \cite{prajwalSubwordLevelLip2022} or hybrid CTC/Attention approaches \cite{maLiRALearningVisual2021, maVisualSpeechRecognition2022}. CTC decoding assumes conditional independence among the output tokens, enabling it to predict the entire token sequence simultaneously \cite{gravesConnectionistTemporalClassification2006}. This results in a trade-off between efficiency and accuracy: while CTC-based models are typically much faster, they tend to be less accurate than autoregressive models, which benefit from the conditional dependency of tokens in the output sequence of an ASR/VSR model \cite{afourasDeepAudioVisualSpeech2022, nozakiRelaxingConditionalIndependence2021}.

Numerous studies present models trained on either unpublished \unpublished~or synthetic datasets \cite{liuSynthVSRScalingVisual2023}, which significantly exceed the volume of the publicly available LRS2 and LRS3 datasets. The use of larger datasets and models has escalated computational demands, with some studies deploying up to 64 industrial-grade GPUs for the training of VSR models \cite{shiLearningAudioVisualSpeech2022,liuSynthVSRScalingVisual2023}. While the use of such resources has indeed advanced state-of-the-art VSR accuracy, it poses significant challenges in energy consumption, resource allocation, and reproducibility, ultimately limiting the practical application of the resultant models.

\presub
\subsection{Contribution}
\postsub

In this work, we present a novel and straightforward method to efficiently train a VSR model by distilling knowledge from a trained speech recognition model. Unlike existing methods that leverage ASR models for VSR model training, our approach directly reuses the top layers of the ASR model, while the lower sections are replicated to infer the intermediate features from visual inputs. We show that our approach yields competitive performance both without using any labeled audio-visual data, and post fine-tuning with a limited quantity of publicly available labeled data. 

To the best of our knowledge, this constitutes the first attempt to train a model capable of performing end-to-end lip-reading without requiring any labeled data during training. 
Further, we investigate the boundaries of this method in terms of computational efficiency by providing a detailed analysis of two of the most influential hyperparameters during training. The outcome is a baseline model that can be trained on a single consumer-grade GPU within a few days and perform real-time inference on a dated CPU.

\presec
\section{Datasets}
\postsec
\vspace{-10pt}
\begin{table}[h]
\centering
\caption{Relevant sentence-level audio-visual datasets.
$^{\dagger}$VoxCeleb2 is not available from the official sources anymore.
$^{\ddagger}$AVSpeech is not available for download and cannot be fully downloaded anymore due to deleted or restricted-access YouTube videos.}
\vspace{5pt}
\label{tab:datasets}
\begin{tabularx}{\columnwidth}{Xrrr}
\toprule
\textbf{Dataset}   & \multicolumn{1}{c}{\textbf{Size {[}hrs{]}}} & \multicolumn{1}{c}{\textbf{Labeled}} & \multicolumn{1}{c}{\textbf{Public}} \\ \midrule
LRS2 (BBC) \cite{chungLipReadingSentences2017} & 224                     & yes              & yes             \\
LRS3 (TED) \cite{afourasLRS3TEDLargescaleDataset2018} & 475                     & yes              & yes             \\ \midrule
VoxCeleb2 \cite{chungVoxCeleb2DeepSpeaker2018}  & 2.4k                    & no               & yes$^{\dagger}$            \\
AVSpeech \cite{ephratLookingListenCocktail2018}   & 4.7k                    & no               & yes$^{\ddagger}$            \\ \midrule
MV-LRS \cite{sonLipReadingProfile2017}     & 775                     & yes              & no              \\
LSVSR \cite{shillingfordLargeScaleVisualSpeech2019}      & 3.8k                    & yes              & no              \\
YT31k \cite{makinoRecurrentNeuralNetwork2019}      & 31k                     & yes              & no              \\
YT90k \cite{serdyukAudioVisualSpeechRecognition2021}      & 90k                     & yes              & no              \\ \bottomrule
\end{tabularx}
\end{table}

Table~\ref{tab:datasets} presents an overview of relevant audio-visual datasets suitable for training a VSR models. LRS2 and LRS3 are the most widely used datasets in related works. The LRS2 dataset encompasses approximately 224 hours of sentence-level videos, extracted from a diverse range of BBC shows. LRS3 contains 475 hours of sentence-level videos sourced from TED and TEDx talks. Despite the existence of larger audio-visual corpora, their lack of public availability limits reproducibility. For this reason, we only use publicly available datasets, i.e., LRS2 and LRS3 datasets for the pre-training and fine-tuning stage, additionally highlighting the data efficiency of our approach. For the unlabeled pre-training, we combine the pre-training sets of LRS2 (195 hrs) and LRS3 (444 hrs), using a total of 639 hours of unlabeled data. The fine-tuning process uses a mix of the LRS2-main (29 hrs) and LRS3-trainval (30 hrs) datasets, making use of 59 hrs of labeled data in total. We evaluate our models on the LRS2 and LRS3 test sets.

\presec
\section{Methodology}
\label{sec:methodology}
\postsec

\begin{figure}[!thb]
\centering 
\includegraphics[width=1\linewidth]{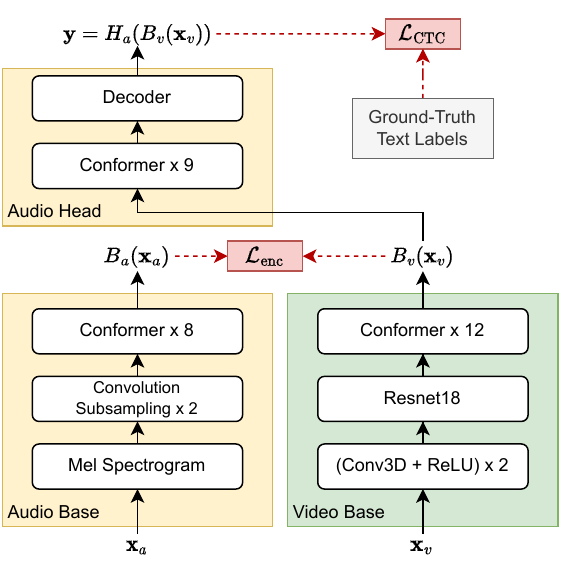}
\vspace{-15pt}
\caption{Architecture of the pre-trained audio model (divided into base and head model $B_a$ and $H_a$) and the visual base model $B_v$.}
\label{fig:architecture} 
\vspace{-10pt}                              
\end{figure}

To distill knowledge from a trained ASR model, our approach divides it into two submodels, the audio base $B_a$ and the audio head $H_a$, as illustrated in Figure~\ref{fig:architecture}. The base submodel transforms the audio signal into a higher-level representation of the speech features present in the audio input. Subsequently, the head part processes these features into a probability distribution of output tokens over time, which is decoded into written text. Our approach involves two steps to train the VSR model: 

\vspace{4pt}
\noindent \textbf{Pre-Training:} We initialize a new base model $B_v$ to handle visual input. The visual base's output aligns in shape with the audio base's output, denoted as $[N, T, d]$, where $N$ is the mini-batch size, $T$ is the length of the longest input sequence in the mini-batch, and $d$ is the encoding dimension of the Conformer layers. The audio base processes the training video's audio, while the visual base handles its corresponding video frames. The visual base is trained to emulate the speech encoding of the audio base, thereby distilling knowledge from the ASR model. In this process, the audio base remains frozen, with training adjustments made solely to the visual base. The combination of the video base and the audio head can then be used to perform end-to-end inference from silent video input to text. 

\vspace{4pt}
\noindent \textbf{Fine-Tuning:} In the fine-tuning stage, we concatenate the visual base from the pre-training step with the audio head. The entire model $H_a(B_v(\mathbf{x}_v))$ is trained end-to-end using the training video's silent video $\mathbf{x}_v$ and ground-truth text labels, employing the same loss function used for the original ASR model (in our case: CTC). Both the visual base and the audio head are subject to training adjustments.

This approach allows the visual base to be trained by using the ASR model's robust speech representations as a target. By utilizing the existing audio head to process the visual base's output, we eli- minate the need for a new head model and, consequently, labeled data during pre-training. As a result, end-to-end VSR becomes feasible right after pre-training using only unlabeled data. While the audio encoding serves as an effective pre-training target for the visual base, its potential is somewhat restricted due to certain acoustic cues missing from the visual input. Through joint training of both the audio head and visual base in the fine-tuning stage, they are adapted to operate together, ultimately enhancing text prediction accuracy. 

\presub
\subsection{Pre-trained ASR Model}
\postsub

For our experiments, we use a pre-trained, Conformer-based ASR model from the Nvidia NeMo toolkit \cite{kuchaievNeMoToolkitBuilding2019}, trained by the CTC loss function. In particular, we chose the \textit{stt\_en\_conformer\_ctc\_small} model with 13M parameters. This model is trained on the \textit{NeMo ASRSet}, a comprehensive ASR corpus encompassing 16,000 hours of transcribed audio from various datasets. 
The model processes audio in the form of a Mel Spectrogram with a window length of 10ms. After a convolutional subsampling stage, the window length of the encodings extends to 40ms, thereby aligning with the frame rate of the video signal at 25 fps. The audio encodings are processed through a sequence of 17 Conformer layers, followed by a linear projection layer that outputs logits for a byte-pair-encoded (BPE) alphabet of size 1025 \cite{sennrichNeuralMachineTranslation2016}. A CTC decoder decodes the sequence of logits into written text. 

The methodology described above can potentially be adapted to any ASR model, regardless of the specific architecture and sequence alignment loss used for its training. We selected this specific model for several reasons: First, it was trained on a vast corpus encompassing various common datasets, and it demonstrates robust performance on challenging ASR benchmarks. Second, it employs a fast CTC decoding scheme and features a simple, yet effective architecture built around a stack of Conformer layers. This design enables easy segmentation into base and head submodels to obtain an intermediate encoding as a target during model pre-training. The model achieves a word error rate (WER) of approximately 10\% when inferring from the audio of the LRS2 validation dataset. This performance is similar to that of the \textit{medium} and \textit{large} variants of the same model. Thus, we opted for the small variant for efficiency reasons.

\presub
\subsection{Model Architecture}
\postsub

The architectures of both the audio and visual models are depicted in Figure~\ref{fig:architecture}. The audio base encompasses the Mel Spectrogram input, convolution subsampling, and the first 8 Conformer layers of the original NeMo toolkit model. The audio head consists of the subsequent 9 Conformer layers, complemented by a linear decoder layer. Conversely, the visual base integrates two 3D convolution layers designed to distill low-level spatio-temporal features from the video input. This is succeeded by a ResNet18 \cite{heDeepResidualLearning2016} that processes each frame individually. To transform the sequence of visual feature vectors into the desired encoding, the baseline model employs 12 Conformer layers and a linear projection layer, which is omitted from the figure for legibility. In contrast to the original audio base from the NeMo model (8 layers with $d=176$) we use 12 layers of $d=256$, as we found this to increase the accuracy of the predictions without significantly increasing the speed of training and inference. 

\presub
\subsection{Data Preparation}
\postsub

To identify the speaker's face in the raw video frames, we use dlib~\cite{kingDlibmlMachineLearning2009} to detect 68 facial landmarks. The center location of the mouth, represented as $(c_{M,x}, c_{M, y})$, is estimated by averaging the coordinates of the mouth's left and right corners, as well as the top and bottom lip centers. Furthermore, the mouth's width $w_M$ is estimated using the Euclidean distance between the mouth's left and right corners. Based on the mouth width $w_M$ we crop the raw frames to a bounding box of $B \times B$ pixels, where $B = bw_M$, and $b \sim \mathcal{U}(2,3)$, representing the Region of Interest (ROI) for lip-reading. The center location for the crop is randomly shifted around the mouth's center location, with a random displacement $d_x, d_y \sim \mathcal{U}(-B/5, B/5)$. Both the random size of the bounding box and the displacement of the center location are used for augmentation purposes and are deactivated during inference/testing.

The resulting crop is normalized, converted to grayscale and resized to the desire input image size. For further image augmentation on the cropped frames, we employ RandAugment \cite{Cubuk_2020_CVPR_Workshops} with parameters \textit{magnitude~=~9} and \textit{num~=~2}. Finally, we use time masking augmentation from \cite{maVisualSpeechRecognition2022} to replace up to 40\% of the video frames by the average frame in every second of the cropped video. 

\presub
\subsection{Model Training}
\postsub
\setlength{\abovedisplayskip}{6pt}
\setlength{\belowdisplayskip}{6pt}

We split each video sample $\mathbf{x}$ into an associated pair of audio data $\mathbf{x}_{a}$ and silent video data $\mathbf{x}_{v}$. For the pre-training phase with unlabeled data we distill knowledge from the audio model by minimizing the mean-squared error between the encodings of the pre-trained audio base $B_a(\mathbf{x}_a)$ feeding the audio data $\mathbf{x}_a$ and the output of the freshly initialized visual base $B_v(\mathbf{x}_v)$ feeding the video data $\mathbf{x}_v$. The encoding loss function $\mathcal{L}_{\text{enc}}$ is defined as
\begin{align}
    \mathcal{L}_{\text{enc}}(\mathbf{x}) = ||B_{a}(\mathbf{x}_{a}) - B_{v}(\mathbf{x}_{v})||_2^2.
\end{align}
Note, that only the visual base $B_v$ is trained in this process, while the audio base $B_a$ is frozen, serving solely as a teacher model. 

For the fine-tuning phase using labeled data we additionally use the CTC loss function find an optimal, monotonic and conditionally independent alignment between the frame-wise predictions and the output sequence. Let $\mathbf{x}_{v}=[x_{v,1}, ..., x_{v,T}]$ be the input frame sequence and $\mathbf{y}=[y_1, ..., y_L]$ the ground-truth text represented as a sequence of target symbols, with $T$ and $L$ as the input and target lengths. With $\mathcal{B}$ mapping a CTC path $\pi$ to an output sequence $y$, the CTC loss is defined as
\setlength{\abovedisplayskip}{3pt}
\setlength{\belowdisplayskip}{3pt}
\begin{align}
    \mathcal{L}_{CTC}(\mathbf{x}_{v},\mathbf{y}) = -\log\sum_{\pi \in \mathcal{B}^{-1}(\mathbf{y})} p(\pi(t)|\mathbf{x}_{v}).
\end{align}
For the pre-training stage, we train the model solely using the encoding loss ($\mathcal{L}_{PT} = \mathcal{L}_\text{enc}$) with unlabeled audio-visual data. Although training only the base part of the network, the resulting model can already be used as an end-to-end VSR model by using $B_v$ in combination with the pre-trained audio head $H_a$ to produce a text output $\mathbf{y} = H_a(B_v(\mathbf{x}_v))$ from silent video input, as shown in Figure~\ref{fig:architecture}.

\setlength{\abovedisplayskip}{6pt}
\setlength{\belowdisplayskip}{6pt}
For fine-tuning the full network, the bottom part of the visual network and the top part of the pre-trained ASR model are jointly trained end-to-end with the CTC loss, while continuing to use the encoding loss as a regularization term:
\begin{align}
    \mathcal{L}_{FT}((\mathbf{x}_{a}, \mathbf{x}_{v},\mathbf{y})) = \mathcal{L}_{CTC}(\mathbf{x}_{v},\mathbf{y}) + \lambda_{\text{enc}}\mathcal{L}_{\text{enc}}(\mathbf{x}),
\end{align}
where we use $\lambda_{\text{enc}} = 1$ for all experiments.

\presub
\subsection{Training Details}
\postsub

We use an Adam Optimizer \cite{adam2014} with $\beta_{1}\!=\!0.9$, $\beta_{2}\!=\!0.98$ and $\epsilon\!=\!10^{-9}$ and a mini-batch size of 16 for all training runs. During the pre-training phase, we maintain a constant learning rate of $1e^{-4}$. For the fine-tuning phase, we linearly warm-up the learning rate for 10,000 steps, peaking at $1e^{-4}$. Post warm-up, the learning rate is proportionally decreased following the inverse square root of the step number \cite{vaswaniAttentionAllYou2017}. Both the pre-training and fine-tuning phase are executed with Automatic Mixed Precision (AMP) \cite{micikeviciusMixedPrecisionTraining2018} allowing for the majority of the model to run on 16-bit floating-point precision. In our case, the use of of AMP does not affect the performance and stability of the models, while it significantly reduces the computational demands of the training process.

\presec
\section{Experimental Evaluation}
\label{sec:results}
\postsec

In this section, we provide a detailed evaluation of the proposed methodology. We begin by examining the performance of a baseline model, reporting word error rates on the LRS2 and LRS3 test datasets after various amounts of pre-training iterations with unsupervised data and, optionally, labeled fine-tuning. We then analyze the outcomes of deviations from this baseline, evaluating the trade-off between model accuracy and training efficiency. Lastly, we present data on the inference speed to highlight the real-time capabilities of our Visual Speech Recognition (VSR) model.
\presub
\subsection{Baseline Results}
\postsub
Two hyperparameters significantly impact the computational resources needed for model training: the duration of the audio/video slices used during pre-training of the visual base, and the size of the input video to the model. For the baseline, we employ video slices of $64\times64$ pixels size and $3$ seconds ($75$ frames) duration, randomly selected from the full video. Figure \ref{fig:res} displays the WER achieved on the LRS2 and LRS3 test sets by the baseline model for a varying number of iterations in the pre-training stage. 
The baseline model is trained on a single \textit{Nvidia RTX3090}, consuming approximately 7GB of GPU memory during pre-training and 12GB during fine-tuning. Pre-training proceeds at a rate of about 10k iterations per hour (4 days per 1M iterations). The fine-tuning process, spanning 20 epochs, requires roughly 12 hours. As a result, our methodology enables the training of a VSR model to achieve a WER of less than 50\% on LRS2 in just 1.5 days (3 days for LRS3) without using industrial-grade deep learning hardware. Our best models achieve a WER of 35\% (LRS2) and 45.7\% (LRS3). When relying solely on unlabeled data, the training duration extends to 16 days to attain a sub-50\% WER on LRS2. The final model can be used to perform real-time inference on a dated laptop CPU (\textit{Intel i7-8550U}), with an average processing time of 250 msec per second of input video.

\input{main_result_figure}
\presub
\subsection{Baseline Variations}
\postsub
Figure \ref{fig:res_hyp} compares different configurations for the input video size~$S$ of the model and the length of video slices $L$ used in pre-training. Naturally, the consumption of computational resources increases monotonically with both $S$ and $L$, while the accuracy plateaus beyond a certain threshold. When comparing the baseline $S=64$ to $S=32$, we observe a marked improvement, but the gains diminish when further increasing the image size to $S=96$. Conversely, this almost doubles the VRAM consumption during both pre-training and fine-tuning. Considering $L$, we can even observe a notable increase in WER for several configurations with $L > 90$. The influence of $L$ on the resource consumption is again substantial, albeit to a lesser degree than $S$.
Therefore, we chose the baseline configuration ($S=64, L=75$) offering a balanced compromise between computational resource demand and prediction accuracy.

\presub
\subsection{Comparison with State-of-the-Art}
\postsub
\input{side_result_figure}
The primary focus of our work is on efficiency and low-resource training using publicly available datasets, rather than achieving another improvement in state-of-the-art VSR performance on LRS2 and LRS3. However, there are noteworthly improvements of our approach as compared to several recent publications in the field. \\
\noindent \textbf{Training on unlabeled data:} To our knowledge, we are the first to report competitive results on the LRS benchmarks (LRS2:~47.4\%, LRS3:~54.7\%) using only unlabeled and publicly available data. Training on 639 hours of entirely unlabeled data, our model is outperforming all methods that utilized public labeled data up until 2020 \cite{petridisAudioVisualSpeechRecognition2018, zhangSpatioTemporalFusionBased2019, afourasASRAllYou2020} as well as several models trained on much larger (unpublished) labeled datasets like LSVSR \cite{shillingfordLargeScaleVisualSpeech2019} and MV-LRS \cite{afourasDeepAudioVisualSpeech2022}.\\
\noindent \textbf{Training with labeled data:} Upon fine-tuning with a limited amount (59 hours) of labeled and publicly available data, we can further reduce the word error rate on these benchmarks (LRS2:~35.0\%, LRS3:~45.7\%). This performance is roughly comparable to that of \cite{maLiRALearningVisual2021} (LRS2:~38.8\%), \cite{maEndToEndAudioVisualSpeech2021} (LRS2:~37.9\%, LRS3:~43.3\%), and \cite{prajwalSubwordLevelLip2022} (LRS2:~28.9\%, LRS3:~40.6\%). Currently, the best-performing model on the LRS benchmarks is trained on 64 Nvidia A100 GPUs using over 3,600 hours of synthetic audio-visual data (LRS3: 43.3\% and 16.9\% with 30 and 3,000 hours of labeled data, respectively) \cite{liuSynthVSRScalingVisual2023}.
In contrast to our approach, all of these methods utilize autoregressive decoders, leading to significantly slower inference times compared to a pure CTC decoding scheme \cite{nozakiRelaxingConditionalIndependence2021}.\\
\noindent \textbf{Other CTC-based VSR models:} Only one model that operates with pure CTC decoding reports better results (LRS3: 38.6\%) \cite{shiLearningAudioVisualSpeech2022} when compared to our approach. This model is trained on 1,759/433 hours of unlabeled/labeled data using 64 Nvidia V100-GPUs, thus relying on substantially more data and higher computational resources. Additionally, \cite{shiLearningAudioVisualSpeech2022} presents results when training their model with limited labeled training data. Using 433 hours of unlabeled data for pre-training, they achieve WERs of 68.8\%, 57.6\%, and 54.2\% on LRS3 with 1, 10, and 100 hours of labeled data, respectively. 
In contrast, our model, pre-trained with 639 hours of unlabeled data, achieves a 54.7\% WER on LRS3 without any labeled data, nearly matching the performance of the model using 100 hours of labeled data in \cite{shiLearningAudioVisualSpeech2022}. After fine-tuning with only 59 hours of labeled data, our model achieves a WER of 45.7\%, thus demonstrating the data efficiency of our approach.

When considering all recent publications that have advanced the state-of-the-art on the LRS benchmarks, our approach most closely aligns with \cite{afourasASRAllYou2020}. This work also employs a pure CTC decoding scheme, utilizes publicly available data, and is trained on reasonable resources, i.e., four GPUs with 11GB of memory each. Despite our reliance on only 7GB of VRAM, we achieve significant improvements over the reported results (LRS2: 51.3\%, LRS3: 59.8\%) in \cite{afourasASRAllYou2020}, even when training exclusively with unlabeled data.

\presec
\section{Conclusion}
\label{sec:conclusion}
\postsec

In this work, we have introduced a novel approach to VSR by learning high-level speech representations from any pre-trained ASR model. By distilling knowledge from the base part and reusing the head part of the ASR model, we facilitate the training of the first VSR model that achieves competitive results on common benchmarks without relying on any labeled or unpublished data. A careful analysis of selected hyperparameters enables us to train our baseline model on consumer hardware and perform real-time inference on a laptop CPU, thereby counteracting the ongoing trend of escalating computational resources in the field.

\bibliographystyle{IEEEbib}
{\footnotesize
\bibliography{VSR_clean}
}

\end{document}

%% file: main_result_figure.tex
\begin{figure}[tb]

\begin{minipage}[b]{.49\linewidth}
  \centering
  \centerline{\includegraphics[width=\linewidth]{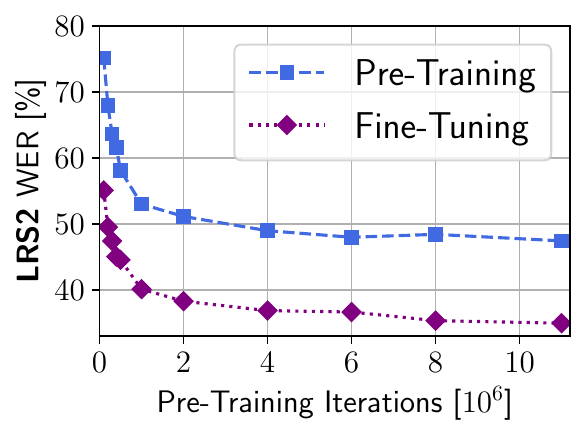}}
\end{minipage}
\hfill
\begin{minipage}[b]{0.49\linewidth}
  \centering
  \centerline{\includegraphics[width=\linewidth]{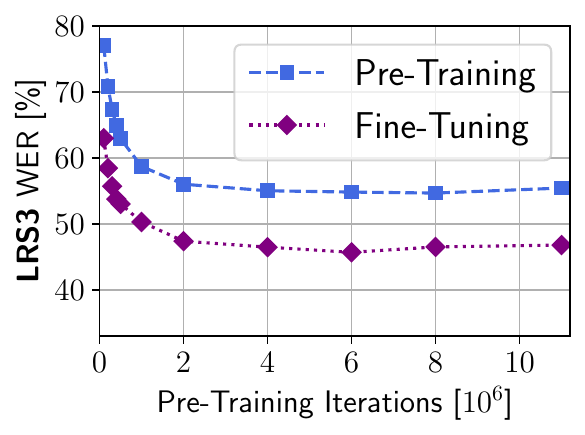}}
\end{minipage}
\vspace{-10pt}
\caption{WER of the baseline model on the LRS2 (left) and LRS3 (right) test sets. The dashed line represents the VSR model's WER after pre-training with unlabeled data for the corresponding number of iterations on the x-axis. The dotted line indicates performance post fine-tuning for 20 epochs (180k iterations) with labeled data, starting from the checkpoint of the specified pre-training iteration.}
\vspace{-10pt}
\label{fig:res}
\end{figure}

%% file: side_result_figure.tex
\begin{figure}[t!]

\begin{minipage}[b]{.49\linewidth}
  \centering
  \centerline{\includegraphics[width=\linewidth]{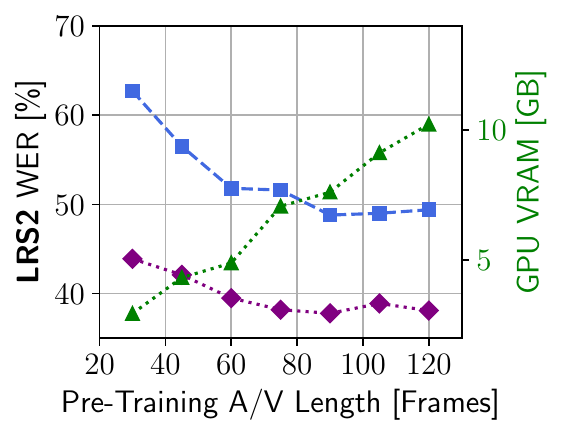}}
\end{minipage}
\hfill
\begin{minipage}[b]{0.49\linewidth}
  \centering
  \centerline{\includegraphics[width=\linewidth]{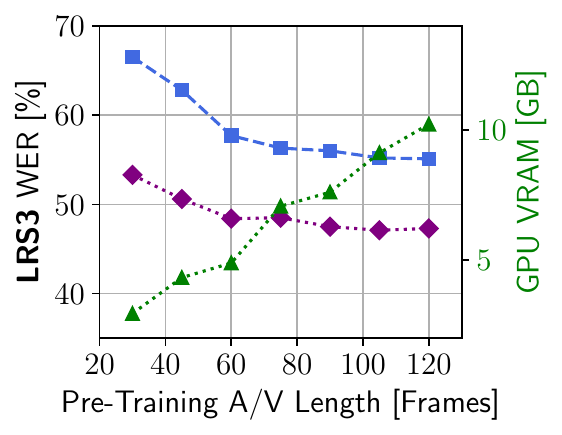}}
\end{minipage}

\begin{minipage}[b]{.49\linewidth}
  \centering
  \centerline{\includegraphics[width=\linewidth]{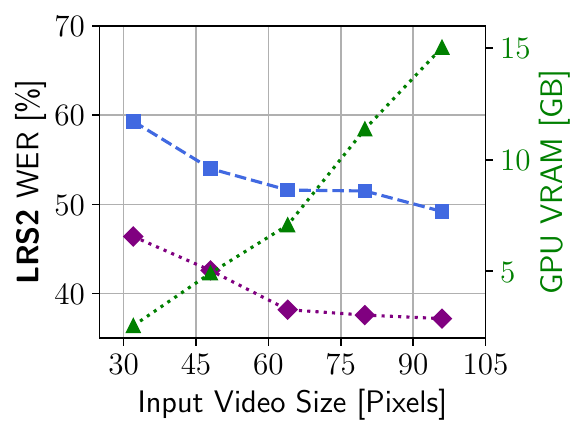}}
\end{minipage}
\hfill
\begin{minipage}[b]{0.49\linewidth}
  \centering
  \centerline{\includegraphics[width=\linewidth]{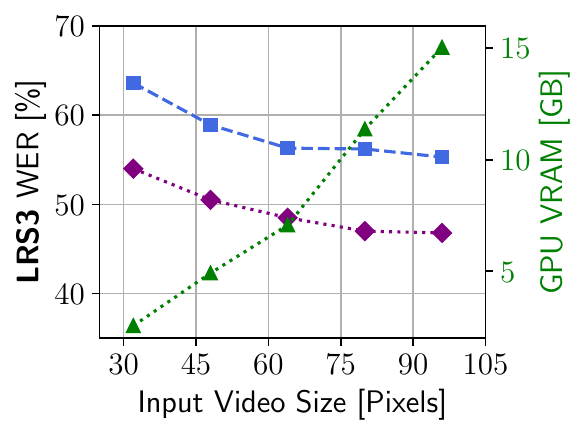}}
\end{minipage}

\vspace{-10pt}
\caption{Results of hyperparameter variation. The top figures show the WER's dependency over the length of the video slice $L$ in frames (during pre-training) for both LRS2 (left) and LRS3 (right). The bottom figures depict the influence of different input video sizes $S$ to the visual base in pixels. The second y-axis shows the video memory consumption for the corresponding pre-training configuration. Each model is trained for 1.7M iterations (150 epochs) in pre-training and, optionally, 90k iterations (20 epochs) in fine-tuning.}
\vspace{-10pt}
\label{fig:res_hyp}
\end{figure}

%% file: main.bbl
\begin{thebibliography}{10}

\bibitem{lauxTwostageVisualSpeech2023}
Hendrik Laux, Ahmed Hallawa, Julio Cesar~Sevarolli Assis, Anke Schmeink, Lukas Martin, and Arne Peine,
\newblock ``Two-stage visual speech recognition for intensive care patients,''
\newblock {\em Scientific Reports}, vol. 13, no. 1, pp. 928, 2023.

\bibitem{serdyukAudioVisualSpeechRecognition2021}
Dmitriy Serdyuk, Otavio Braga, and Olivier Siohan,
\newblock ``Audio-{{Visual Speech Recognition}} is {{Worth}} \$32\textbackslash times 32\textbackslash times 8\$ {{Voxels}},''
\newblock in {\em {{IEEE Automatic Speech Recognition}} and {{Understanding Workshop}} ({{ASRU}})}. 2021, pp. 796--802, {IEEE}.

\bibitem{xuDiscriminativeMultiModalitySpeech2020}
Bo~Xu, Cheng Lu, Yandong Guo, and Jacob Wang,
\newblock ``Discriminative {{Multi-Modality Speech Recognition}},''
\newblock in {\em Proceedings of the {{IEEE}}/{{CVF Conference}} on {{Computer Vision}} and {{Pattern Recognition}} ({{CVPR}})}. 2020, pp. 14433--14442, {IEEE}.

\bibitem{goldschenRationalePhonemevisemeMapping1996}
Alan~J. Goldschen, Oscar~N. Garcia, and Eric~D. Petajan,
\newblock ``Rationale for phoneme-viseme mapping and feature selection in visual speech recognition,''
\newblock in {\em Speechreading by Humans and Machines: {{Models}}, Systems, and Applications}, pp. 505--515. {Springer Berlin Heidelberg}, 1996.

\bibitem{chungLipReadingSentences2017}
Joon~Son Chung, Andrew Senior, Oriol Vinyals, and Andrew Zisserman,
\newblock ``Lip {{Reading Sentences}} in the {{Wild}},''
\newblock in {\em Proceedings of the {{IEEE}}/{{CVF Conference}} on {{Computer Vision}} and {{Pattern Recognition}} ({{CVPR}})}. 2017, pp. 3444--3453, {IEEE}.

\bibitem{shillingfordLargeScaleVisualSpeech2019}
Brendan Shillingford, Yannis Assael, Matthew~W. Hoffman, Thomas Paine, C{\'i}an Hughes, Utsav Prabhu, et~al.,
\newblock ``Large-{{Scale Visual Speech Recognition}},''
\newblock in {\em Interspeech}. 2019, pp. 4135--4139, {ISCA}.

\bibitem{shiLearningAudioVisualSpeech2022}
Bowen Shi, Wei-Ning Hsu, Kushal Lakhotia, and Abdelrahman Mohamed,
\newblock ``Learning {{Audio-Visual Speech Representation}} by {{Masked Multimodal Cluster Prediction}},''
\newblock {\em arXiv preprint arXiv:2201.02184}, 2022.

\bibitem{liuSynthVSRScalingVisual2023}
Xubo Liu, Egor Lakomkin, Konstantinos Vougioukas, Pingchuan Ma, Honglie Chen, Ruiming Xie, et~al.,
\newblock ``{{SynthVSR}}: {{Scaling Up Visual Speech Recognition With Synthetic Supervision}},''
\newblock in {\em Proceedings of the {{IEEE}}/{{CVF Conference}} on {{Computer Vision}} and {{Pattern Recognition}} ({{CVPR}})}. 2023, pp. 18806--18815, {IEEE}.

\bibitem{prajwalSubwordLevelLip2022}
K~R Prajwal, Triantafyllos Afouras, and Andrew Zisserman,
\newblock ``Sub-word {{Level Lip Reading With Visual Attention}},''
\newblock in {\em Proceedings of the {{IEEE}}/{{CVF Conference}} on {{Computer Vision}} and {{Pattern Recognition}} ({{CVPR}})}. 2022, pp. 5152--5162, {IEEE}.

\bibitem{sonLipReadingProfile2017}
Joon~Son Son and Andrew Zisserman,
\newblock ``Lip {{Reading}} in {{Profile}},''
\newblock in {\em Procedings of the {{British Machine Vision Conference}}}. 2017, p. 155, {British Machine Vision Association}.

\bibitem{makinoRecurrentNeuralNetwork2019}
Takaki Makino, Hank Liao, Yannis Assael, Brendan Shillingford, Basilio Garcia, Otavio Braga, et~al.,
\newblock ``Recurrent {{Neural Network Transducer}} for {{Audio-Visual Speech Recognition}},''
\newblock in {\em {{IEEE Automatic Speech Recognition}} and {{Understanding Workshop}} ({{ASRU}})}. 2019, pp. 905--912, {IEEE}.

\bibitem{assaelLipNetEndtoEndSentencelevel2016}
Yannis~M. Assael, Brendan Shillingford, Shimon Whiteson, and Nando {de Freitas},
\newblock ``{{LipNet}}: {{End-to-End Sentence-level Lipreading}},''
\newblock {\em arXiv preprint arXiv:1611.01599}, 2016.

\bibitem{gravesConnectionistTemporalClassification2006}
Alex Graves, Santiago Fernandez, Faustino Gomez, and Jurgen Schmidhuber,
\newblock ``Connectionist {{Temporal Classification}}: {{Labelling Unsegmented Sequence Data}} with {{Recurrent Neural Networks}},''
\newblock in {\em International {{Conference}} on {{Machine Learning}} ({{ICML}})}. 2006, pp. 369--376, {ACM Press}.

\bibitem{cookeAudiovisualCorpusSpeech2006}
Martin Cooke, Jon Barker, Stuart Cunningham, and Xu~Shao,
\newblock ``An audio-visual corpus for speech perception and automatic speech recognition,''
\newblock {\em The Journal of the Acoustical Society of America}, vol. 120, no. 5, pp. 2421--2424, 2006.

\bibitem{afourasLRS3TEDLargescaleDataset2018}
Triantafyllos Afouras, Joon~Son Chung, and Andrew Zisserman,
\newblock ``{{LRS3-TED}}: A large-scale dataset for visual speech recognition,''
\newblock {\em arXiv preprint arXiv:1809.00496}, 2018.

\bibitem{vaswaniAttentionAllYou2017}
Ashish Vaswani, Noam Shazeer, Niki Parmar, Jakob Uszkoreit, Llion Jones, Aidan~N. Gomez, et~al.,
\newblock ``Attention is {{All}} you {{Need}},''
\newblock in {\em Proceedings of the 31st {{International Conference}} on {{Neural Information Processing Systems}} ({{NIPS}})}. 2017, {{NIPS}}'17, pp. 6000--6010, {Curran Associates Inc.}

\bibitem{gulatiConformerConvolutionaugmentedTransformer2020}
Anmol Gulati, James Qin, Chung-Cheng Chiu, Niki Parmar, Yu~Zhang, Jiahui Yu, et~al.,
\newblock ``Conformer: {{Convolution-augmented Transformer}} for {{Speech Recognition}},''
\newblock in {\em Interspeech}. 2020, pp. 5036--5040, {ISCA}.

\bibitem{maEndToEndAudioVisualSpeech2021}
Pingchuan Ma, Stavros Petridis, and Maja Pantic,
\newblock ``End-{{To-End Audio-Visual Speech Recognition}} with {{Conformers}},''
\newblock in {\em {{IEEE International Conference}} on {{Acoustics}}, {{Speech}} and {{Signal Processing}} ({{ICASSP}})}. 2021, pp. 7613--7617, {IEEE}.

\bibitem{afourasASRAllYou2020}
Triantafyllos Afouras, Joon~Son Chung, and Andrew Zisserman,
\newblock ``{{ASR}} is {{All You Need}}: {{Cross-Modal Distillation}} for {{Lip Reading}},''
\newblock in {\em {{IEEE International Conference}} on {{Acoustics}}, {{Speech}} and {{Signal Processing}} ({{ICASSP}})}. 2020, pp. 2143--2147, {IEEE}.

\bibitem{maLiRALearningVisual2021}
Pingchuan Ma, Rodrigo Mira, Stavros Petridis, Bj{\"o}rn~W. Schuller, and Maja Pantic,
\newblock ``{{LiRA}}: {{Learning Visual Speech Representations}} from {{Audio Through Self-Supervision}},''
\newblock in {\em Interspeech}. 2021, pp. 3011--3015, {ISCA}.

\bibitem{maVisualSpeechRecognition2022}
Pingchuan Ma, Stavros Petridis, and Maja Pantic,
\newblock ``Visual {{Speech Recognition}} for {{Multiple Languages}} in the {{Wild}},''
\newblock {\em Nature Machine Intelligence}, vol. 4, no. 11, pp. 930--939, 2022.

\bibitem{afourasDeepAudioVisualSpeech2022}
Triantafyllos Afouras, Joon~Son Chung, Andrew Senior, Oriol Vinyals, and Andrew Zisserman,
\newblock ``Deep {{Audio-Visual Speech Recognition}},''
\newblock {\em IEEE Transactions on Pattern Analysis and Machine Intelligence}, vol. 44, no. 12, pp. 8717--8727, 2022.

\bibitem{nozakiRelaxingConditionalIndependence2021}
Jumon Nozaki and Tatsuya Komatsu,
\newblock ``Relaxing the {{Conditional Independence Assumption}} of {{CTC-Based ASR}} by {{Conditioning}} on {{Intermediate Predictions}},''
\newblock in {\em Interspeech}. 2021, pp. 3735--3739, {ISCA}.

\bibitem{chungVoxCeleb2DeepSpeaker2018}
Joon~Son Chung, Arsha Nagrani, and Andrew Zisserman,
\newblock ``{{VoxCeleb2}}: {{Deep Speaker Recognition}},''
\newblock in {\em Interspeech}. 2018, pp. 1086--1090, {ISCA}.

\bibitem{ephratLookingListenCocktail2018}
Ariel Ephrat, Inbar Mosseri, Oran Lang, Tali Dekel, Kevin Wilson, Avinatan Hassidim, et~al.,
\newblock ``Looking to listen at the cocktail party: A speaker-independent audio-visual model for speech separation,''
\newblock {\em ACM Transactions on Graphics}, vol. 37, no. 4, pp. 1--11, 2018.

\bibitem{kuchaievNeMoToolkitBuilding2019}
Oleksii Kuchaiev, Jason Li, Huyen Nguyen, Oleksii Hrinchuk, Ryan Leary, Boris Ginsburg, et~al.,
\newblock ``{{NeMo}}: A toolkit for building {{AI}} applications using {{Neural Modules}},''
\newblock {\em arXiv preprint arXiv:1909.09577}, 2019.

\bibitem{sennrichNeuralMachineTranslation2016}
Rico Sennrich, Barry Haddow, and Alexandra Birch,
\newblock ``Neural {{Machine Translation}} of {{Rare Words}} with {{Subword Units}},''
\newblock in {\em Proceedings of the 54th {{Annual Meeting}} of the {{Association}} for {{Computational Linguistics}}}. 2016, pp. 1715--1725, {Association for Computational Linguistics}.

\bibitem{heDeepResidualLearning2016}
Kaiming He, Xiangyu Zhang, Shaoqing Ren, and Jian Sun,
\newblock ``Deep {{Residual Learning}} for {{Image Recognition}},''
\newblock in {\em Proceedings of the {{IEEE}}/{{CVF Conference}} on {{Computer Vision}} and {{Pattern Recognition}} ({{CVPR}})}. 2016, pp. 770--778, {IEEE}.

\bibitem{kingDlibmlMachineLearning2009}
Davis~E King,
\newblock ``Dlib-ml: {{A Machine Learning Toolkit}},''
\newblock {\em The Journal of Machine Learning Research}, vol. 10, pp. 1755--1758, 2009.

\bibitem{Cubuk_2020_CVPR_Workshops}
Ekin~D. Cubuk, Barret Zoph, Jonathon Shlens, and Quoc~V. Le,
\newblock ``Randaugment: {{Practical}} automated data augmentation with a reduced search space,''
\newblock in {\em Proceedings of the {{IEEE}}/{{CVF Conference}} on {{Computer Vision}} and {{Pattern Recognition}} ({{CVPR}}) {{Workshops}}}. 2020, pp. 702--703, {IEEE}.

\bibitem{adam2014}
Diederik Kingma and Jimmy Ba,
\newblock ``Adam: {{A Method}} for {{Stochastic Optimization}},''
\newblock in {\em Proceedings of the {{International Conference}} on {{Learning Representations}} ({{ICLR}})}. 2015, {IEEE}.

\bibitem{micikeviciusMixedPrecisionTraining2018}
Paulius Micikevicius, Sharan Narang, Jonah Alben, Gregory Diamos, Erich Elsen, David Garcia, et~al.,
\newblock ``Mixed {{Precision Training}},''
\newblock in {\em Proceedings of the {{International Conference}} on {{Learning Representations}} ({{ICLR}})}. 2018, {IEEE}.

\bibitem{petridisAudioVisualSpeechRecognition2018}
Stavros Petridis, Themos Stafylakis, Pingchuan Ma, Georgios Tzimiropoulos, and Maja Pantic,
\newblock ``Audio-{{Visual Speech Recognition}} with a {{Hybrid CTC}}/{{Attention Architecture}},''
\newblock in {\em {{IEEE Spoken Language Technology Workshop}} ({{SLT}})}. 2018, pp. 513--520, {IEEE}.

\bibitem{zhangSpatioTemporalFusionBased2019}
Xingxuan Zhang, Feng Cheng, and Wang Shilin,
\newblock ``Spatio-{{Temporal Fusion Based Convolutional Sequence Learning}} for {{Lip Reading}},''
\newblock in {\em {{IEEE}}/{{CVF International Conference}} on {{Computer Vision}} ({{ICCV}})}. 2019, pp. 713--722, {IEEE}.

\end{thebibliography}
